\documentclass[letterpaper,10pt,conference]{ieeeconf}

\usepackage{cite}
\usepackage{amsmath,amssymb,amsfonts}
\usepackage{graphicx}
\usepackage{hyperref}
\usepackage{textcomp}
\usepackage{tabularx}
\usepackage{balance}
\usepackage{multirow}
\usepackage{dblfloatfix}
\usepackage{siunitx}
\usepackage{booktabs}
\usepackage{longtable}
\usepackage{color,colortbl}
\usepackage{algpseudocode}
\usepackage{algorithm}
\usepackage{xcolor}
\usepackage{float}
\def\BibTeX{{\rm B\kern-.05em{\sc i\kern-.025em b}\kern-.08em
    T\kern-.1667em\lower.7ex\hbox{E}\kern-.125emX}}
\definecolor{Gray2}{gray}{0.9}
\definecolor{Gray}{gray}{0.7}

\IEEEoverridecommandlockouts
\begin{document}

% \end{document}
% \title{\LARGE \bf
% GoalSwarm: Multi-UAV Semantic Coordination for Open-Vocabulary Object Navigation
% }

% \begin{document}
% \maketitle

% \title{\LARGE \bf AgilePilot: Intelligent Drone Motion Planning with Deep Reinforcement Learning in Dynamic Environments Leveraging Real-time Object Detection\\
% }
\title{\LARGE \bf GoalSwarm: Multi-UAV Semantic Coordination for Open-Vocabulary
Object Navigation\\
}

\author{MoniJesu James, Amir Atef Habel, Aleksey Fedoseev, and Dzmitry Tsetserokou
\thanks{The authors are with the Intelligent Space Robotics Laboratory, Center for Digital Engineering, Skolkovo Institute of Science and Technology, Moscow, Russia. 
\tt \{monijesu.james, amir.habel, aleksey.fedoseev, d.tsetserokou\}@skoltech.ru}
% \thanks{Research reported in this publication was financially supported by the RSF-DST grant No. 24-41-02039.}
}
\maketitle

\begin{abstract}
Cooperative visual semantic navigation is a foundational capability for aerial robot teams operating in unknown environments. However, achieving robust open-vocabulary object-goal navigation remains challenging due to the computational constraints of deploying heavy perception models onboard and the complexity of decentralized multi-agent coordination. We present \textit{GoalSwarm}, a fully decentralized multi-UAV framework for zero-shot semantic object-goal navigation. Each UAV collaboratively constructs a shared, lightweight 2D top-down semantic occupancy map by projecting depth observations from aerial vantage points, eliminating the computational burden of full 3D representations while preserving essential geometric and semantic structure.
The core contributions of GoalSwarm are threefold: (1)~integration of zero-shot foundation model --- SAM3 for open vocabulary detection and pixel-level segmentation, enabling open-vocabulary target identification without task-specific training; (2)~a Bayesian Value Map that fuses multi-viewpoint detection confidences into a per-pixel goal-relevance distribution, enabling informed frontier scoring via Upper Confidence Bound (UCB) exploration; and (3)~a decentralized coordination strategy combining semantic frontier extraction, cost-utility bidding with geodesic path costs, and spatial separation penalties to minimize redundant exploration across the swarm.
 \end{abstract}


\section{Introduction}

Autonomous multi-agent drone swarms are increasingly applied in search and rescue, environmental monitoring, and infrastructure inspection, where the success of the mission relies on a team's ability to collaboratively explore unknown environments and locate targets. This requires both efficient exploration and semantic understanding under limited onboard resources.

A number of researches explore the ability of the robot swarms to navigate efficiently by relying on modular pipelines that combine mapping and planning. For instance, SLAM-based navigation with robot swarm proposed by Friess et al. ~\cite{Friess_2024} enables distributed mapping, while graph-based methods by Walker et al. ~\cite{Walker_2020} coordinate exploration in partially observable environments. However, such methods rely on a complex preprocessing of the drone observation data and multiple bulky sensors, limiting the payload capacity of the drone. Several works aim to achieve semantic mapping with 3D neural reconstruction, e.g., NeuroSwarm by Zhura et al. ~\cite{Zhura_2023}. While achieving high reliability and planning optimization, these approaches are computationally demanding, thus limiting the speed of the navigation.
The above-mentioned methods either lack semantic flexibility or are computationally expensive, making their application limited for drones. To bypass these limitations in the Object-goal Navigation (ObjectNav) problem, we propose GoalSwarm, a decentralized multi-UAV framework for zero-shot ObjectNav. The developed pipeline builds a lightweight 2D semantic map instead of a full 3D reconstruction of the environment and integrates foundation models with a Bayesian Value Map for efficient, uncertainty-aware exploration, enabling scalable and robust multi-agent target navigation.
\begin{figure}[t!]
    \centering
    \includegraphics[width=0.8\linewidth, alt={Real flight experiment of Media Pipe and A* planner ensuring safe handover position of the drone with respect to the human}]{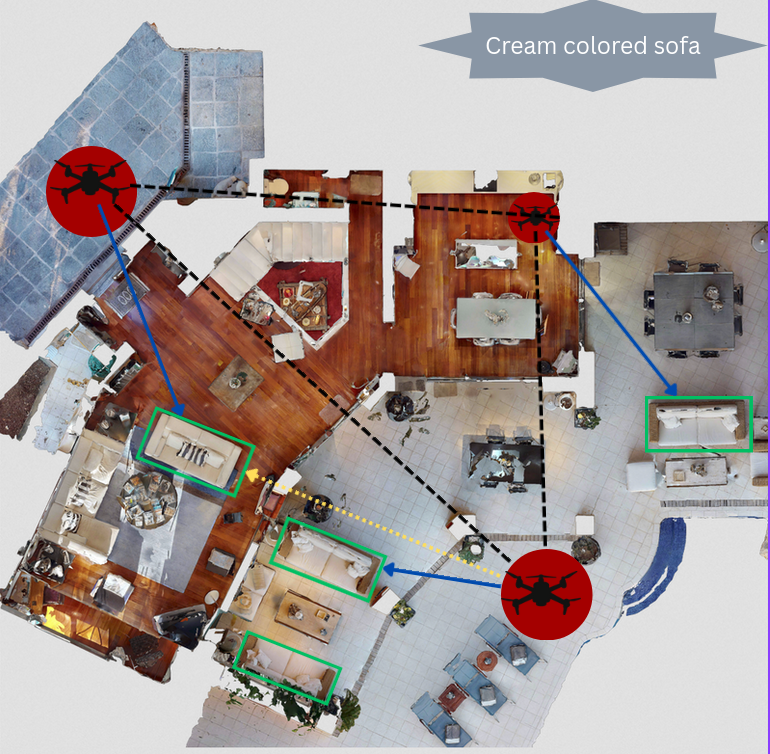}
    \caption{Open vocabulary semantic coordination for multi-UAV object-goal navigation}
    \label{fig:goalswarm}
\end{figure}

\section{Related Work}

\subsection{Multi-Robot Object-Goal Navigation}

ObjectNav requires an agent to locate and navigate toward a specific object in an unseen environment. Single-agent approaches have leveraged modular pipelines combining semantic mapping with frontier-based exploration ~\cite{chaplot2020object, ramakrishnan2022poni}.
Early end-to-end frameworks focused on improving visual representation by combining RGB-D images with semantic segmentation ~\cite{mousavian_2019} and graph-encoded relational information about objects ~\cite{Du_2020}, achieving significant improvements in scene understanding. More recent approaches additionally target the mapping of visual data to actions through reinforcement learning ~\cite{Khan_2025} or VLM models ~\cite{Hu_2025}.

Multi-robot extensions introduce the challenge of coordination: CoNAV~\cite{jain2019two} and Co-NavGPT~\cite{yu2023co} explored cooperative strategies using shared maps and LLM-based task allocation, respectively. However, these methods rely on ground-robot assumptions (navmesh-based planning) and closed-vocabulary detection, limiting their applicability to aerial platforms and novel object categories.

\subsection{Vision-Language Models for Navigation}
Recent advances in vision-language foundation models have enabled open-vocabulary scene understanding. GroundingDINO~\cite{liu2024grounding} combines DINO with grounded pre-training for open-set object detection, while the Segment Anything Model (SAM)~\cite{kirillov2023segment} provides class-agnostic segmentation, SAM3 ~\cite{carion2025sam} provides a single model for open vocabulary detection, segmentation, and tracking of any object in any given image or video. VLFM~\cite{yokoyama2024vlfm} demonstrated value maps for single-robot navigation using BLIP-2 confidence scores. Our work extends this paradigm to multi-UAV systems by replacing single-agent VLM queries with a decentralized Bayesian fusion scheme across multiple viewpoints.

\subsection{Aerial Semantic Navigation}
UAV navigation in indoor environments presents unique challenges compared to ground robots: full 3D freedom of movement, altitude-dependent perception, and the absence of navigable surface meshes. Some works explore a centralized approach for swarm VLM-based scenery understanding ~\cite{Zafar_2025}, allowing usage of a single camera with a top-down view that provides comprehensive cues of goals and obstacles. VisFly~\cite{li2025visfly} provides a simulation framework for vision-based drone navigation in photorealistic scenes. To our knowledge, GoalSwarm is the first framework to combine multi-UAV coordination, open-vocabulary grounding, and Bayesian spatial reasoning for aerial object-goal navigation.

\section{Methodology}

Our proposed framework, \textit{GoalSwarm}, coordinates a swarm of $N$ UAVs to achieve open-vocabulary object-goal navigation in unknown, photorealistic indoor environments. The system is fully decentralized: each UAV maintains a local perception and mapping pipeline, periodically synchronizes with the swarm, and independently selects exploration targets. The framework comprises four primary modules described below.

\begin{figure*}[t!]
    \centering
    \includegraphics[width=0.8\linewidth]{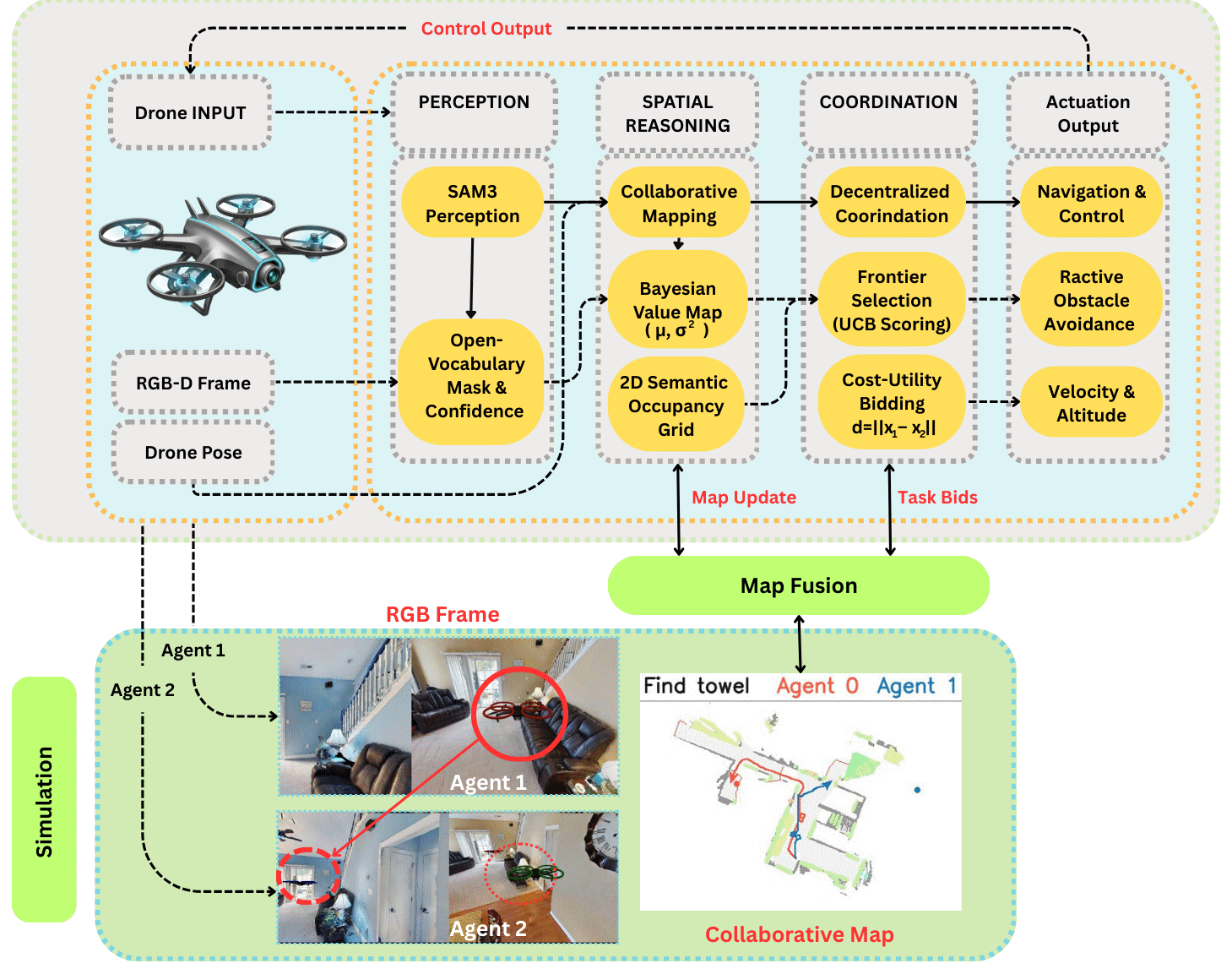}
    \caption{Multiple agents process vision-language cues, perform local planning, and share semantic maps.}
    \label{fig:architecture}
\end{figure*}

\subsection{System Overview}
At each timestep $t$, UAV $j \in \{1, \ldots, N\}$ executes the following loop:
\begin{enumerate}
    \item \textbf{Observe}: Acquire RGB-D frame $(\mathbf{I}_t^j, \mathbf{D}_t^j)$ and odometry pose $\mathbf{p}_t^j = (x, y, z, \theta)$.
    \item \textbf{Perceive}: Run zero-shot grounding to detect target objects and compute a detection confidence $v_t^j \in [0, 1]$.
    \item \textbf{Map}: Project depth into a 2D top-down semantic occupancy grid and update the Bayesian value map with $v_t^j$.
    \item \textbf{Coordinate}: Extract frontiers from the shared map, score them via UCB, and select the optimal frontier.
    \item \textbf{Navigate}: Execute depth-based reactive navigation toward the selected waypoint using a PID controller.
\end{enumerate}

\subsection{Collaborative Semantic Mapping}
Each UAV constructs and maintains a 2D top-down semantic occupancy map $\mathcal{M} \in \mathbb{R}^{W \times W \times (2 + K)}$, where $W = \lfloor L / r \rfloor$ is the map width, $L$ is the physical map extent (24\,m), $r = 5$\,cm is the cell resolution, and $K = 16$ is the number of semantic categories.

\subsubsection{Depth-to-Voxel Projection}
Given a depth image $\mathbf{D}_t \in \mathbb{R}^{H \times W_f}$ from a camera with an intrinsic matrix $\mathbf{K}$ and horizontal field of view $\phi$, we first compute the 3D point cloud in the camera frame:
\begin{equation}
    \mathbf{X}_c = \frac{(u - c_x) \cdot \mathbf{D}(u,v)}{f}, \quad
    \mathbf{Z}_c = \frac{(v - c_v) \cdot \mathbf{D}(u,v)}{f}
\end{equation}
where $f = \frac{W_f}{2 \tan(\phi/2)}$ is the focal length and $(c_x, c_v)$ is the principal point. The point cloud is transformed into the geocentric frame by applying a rotation $\mathbf{R}(\alpha)$ for the camera elevation angle $\alpha$ and translating by the sensor height $h_s$:
\begin{equation}
    \mathbf{P}_{geo} = \mathbf{R}(\alpha) \cdot \mathbf{P}_{cam} + [0, 0, h_s]^\top
\end{equation}

Points are then voxelized into a 3D grid with $xy$-resolution $r$ and $z$-resolution $r_z$. The voxel grid is collapsed along the height axis within a valid range $[z_{min}, z_{max}]$ relative to the agent height $h_a$ to produce a 2D occupancy prediction:
\begin{equation}
    \hat{m}_{obs}(i,j) = \mathbb{1}\left[\sum_{k=k_{min}}^{k_{max}} \text{voxel}(i,j,k) > \tau_{map}\right]
\end{equation}
where $k_{min} = \lfloor 25\text{cm} / r_z \rfloor$ and $k_{max} = \lfloor (h_a + 50\text{cm}) / r_z \rfloor$ define the obstacle height band, and $\tau_{map}$ is the occupancy threshold.

\subsubsection{Map Fusion}
The local agent-view map is transformed to the global frame via affine grid sampling using the agent's cumulative pose. The global map is updated by element-wise max fusion:
\begin{equation}
    \mathcal{M}_{t+1} = \max(\mathcal{M}_t, \mathcal{T}(\hat{\mathcal{M}}_t^{local}; \mathbf{p}_t))
\end{equation}
where $\mathcal{T}(\cdot; \mathbf{p}_t)$ denotes the spatial transformer parameterized by the agent's pose.

\subsubsection{Multi-Agent Asynchronous Synchronization}
In the multi-UAV setting, each agent maintains its own local map instance. Maps are periodically broadcast and fused into a shared global map using the same max-fusion operator. Because the map is a lightweight 2D grid ($W \times W \times (2+K)$ with $W = 480$), the communication overhead per synchronization is approximately $480 \times 480 \times 18 \times 4\text{B} \approx 6.6$\,MB uncompressed, which can be further reduced via delta encoding.

\subsection{Zero-Shot Object and Language Goal Grounding}
GoalSwarm eliminates reliance on closed-vocabulary detectors by leveraging zero-shot foundation models for open-vocabulary target identification.

\subsubsection{Detection Pipeline}
Given an open-vocabulary target description $g$ (e.g., ``red chair'') and an RGB frame $\mathbf{I}_t$, we employ SAM3, a direct single model for both open vocabulary detection, segmentation, and tracking.
The aggregated detection confidence is:
\begin{equation}
    v_t = \max_i s_i \cdot \frac{|\mathbf{m}_i|}{H \cdot W_f}
\end{equation}
combining the detector's confidence with the relative mask area to down-weight spurious small detections.

\subsubsection{Bayesian Value Map}
Detection confidences are spatially anchored via the \emph{Bayesian Value Map} (BVM), a per-pixel probabilistic map tracking goal relevance. Each cell stores a mean $\mu$ and variance $\sigma^2$. Given a new observation with mean $\mu_{obs}$ and variance $\sigma^2_{obs}$ (derived from the depth-based viewing confidence cone), the posterior is updated via Bayesian fusion:
\begin{equation}
    \mu_{t+1} = \frac{\sigma_{obs}^2 \, \mu_t + \sigma_t^2 \, \mu_{obs}}{\sigma_t^2 + \sigma_{obs}^2 + \epsilon}
\end{equation}
\begin{equation}
    \sigma_{t+1}^2 = \frac{\sigma_t^2 \, \sigma_{obs}^2}{\sigma_t^2 + \sigma_{obs}^2 + \epsilon}
\end{equation}
where $\epsilon \to 0^+$ is a regularization constant. The observation variance $\sigma^2_{obs} = 1 - c(\mathbf{x})$ is derived from a depth-dependent confidence cone $c(\mathbf{x}) \in [c_{min}, 1]$ that attenuates with distance and angular offset from the camera optical axis:
\begin{equation}
    c(\mathbf{x}) = \cos^2\!\left(\frac{\theta(\mathbf{x})}{\phi / 2} \cdot \frac{\pi}{2}\right) \cdot \text{remap}(c_{min}, 1)
\end{equation}
where $\theta(\mathbf{x})$ is the angle from the camera center to pixel $\mathbf{x}$ and $c_{min} = 0.25$ is the minimum edge confidence.

\subsection{Decentralized Multi-UAV Coordination}

\subsubsection{Semantic Frontier Extraction}
Frontiers (boundaries between explored free space and unknown territory) are extracted from the shared global map by detecting cells where explored area ($\mathcal{M}[:,:,1] > 0$) is adjacent to unexplored cells. Each frontier cluster $f_i$ is scored using the Upper Confidence Bound (UCB) derived from the BVM:
\begin{equation}
    U(f_i) = \tilde{\mu}(f_i) + \beta \sqrt{\max(0, \, \tilde{\sigma}^2(f_i))}
\end{equation}
where $\tilde{\mu}(f_i)$ and $\tilde{\sigma}^2(f_i)$ are the median mean and variance over a circular footprint of radius $R$ around the frontier centroid, and $\beta = 1.7$ controls the exploration-exploitation trade-off.

\subsubsection{Cost-Utility Bidding}
Multi-UAV frontier assignment is formulated as a distributed cost-utility optimization. For UAV $j$ at position $\mathbf{p}_j$ and frontier $f_i$, the composite score is:
\begin{equation}
    \text{Score}_{j,i} = \omega_1 U(f_i) - \omega_2 C(\mathbf{p}_j, f_i) + \omega_3 S(f_i) - \mathcal{P}_{sep}
\end{equation}
where $C(\mathbf{p}_j, f_i)$ is the geodesic path cost computed via the Fast Marching Method (FMM) on the occupancy grid, $S(f_i)$ is the frontier size (number of boundary cells), and $\mathcal{P}_{sep}$ is a spatial separation penalty enforcing minimum inter-UAV distance:
\begin{equation}
    \mathcal{P}_{sep} = \lambda \cdot \max\!\left(0, \, d_{min} - \min_{k \neq j} \|\mathbf{p}_k - f_i\|\right)
\end{equation}

\subsubsection{Conflict Resolution and Target Pursuit}
UAVs asynchronously broadcast their selected frontiers. Conflicts (two UAVs targeting the same frontier) are resolved by a priority scheme favoring the UAV with the highest score. Once a target is positively localized with high confidence ($\tilde{\mu} > \tau_{goal}$), the nearest UAV transitions from exploration to exploitation mode, navigating directly to the target coordinates.

\subsection{UAV Navigation Controller}
Unlike ground robots that rely on navigable surface meshes, our UAVs use a reactive depth-based navigation controller with full 3D freedom of movement.

\subsubsection{Depth-Based Obstacle Avoidance}
The depth image $\mathbf{D}_t$ is divided into five angular sectors (left, slight-left, center, slight-right, right). For each sector $s$, we compute the mean obstacle distance $\bar{d}_s$. An obstacle is detected when $\bar{d}_s < d_{safe}$, where $d_{safe} = 1.0$\,m is the safety threshold. The navigation action is selected by a priority cascade:
\begin{enumerate}
    \item If \textbf{stuck} (position unchanged for $> 10$ steps): execute a random escape maneuver (backward + random turn).
    \item If \textbf{center blocked}: turn toward the direction with maximum clearance.
    \item If \textbf{goal found} ($v_t > \tau_{goal}$): move directly forward toward the target.
    \item Otherwise: turn toward the relative angle to the selected frontier and move forward.
\end{enumerate}

\subsubsection{Altitude Management}
UAVs operate at fixed cruise altitudes that vary by operational phase: high altitude ($h = 3.0$\,m) for wide-area survey, medium ($h = 2.0$\,m) for room-level search, and low ($h = 1.5$\,m) for close-range object inspection. Altitude transitions are controlled by a PID controller.

\section{Experimental Setup}

\subsection{Simulation Environment}
We evaluate GoalSwarm on the GOAT-Bench~\cite{khanna2024goatbenchbenchmarkmultimodallifelong} benchmark, which defines sequential multi-object navigation episodes within the Habitat simulator~\cite{savva2019habitat} using photorealistic HM3D scenes~\cite{yokoyama2024hm3d}. Each episode specifies a sequence of target objects; we evaluate on the \emph{object-type} subtasks (open-vocabulary ObjectNav) and skip description-type subtasks. For aerial dynamics, we additionally leverage the VisFly~\cite{li2025visfly} framework, which provides realistic quadrotor physics including collision detection via depth-based KD-tree proximity queries. Each UAV is modeled as a compact indoor aerial platform (height 1.41\,m, radius 0.17\,m) equipped with a front-facing RGB-D camera at 1.31\,m height ($360 \times 640$ resolution, $42^\circ$ HFOV, depth range 0.5--5.0\,m). Agents execute six discrete actions: \texttt{STOP}, \texttt{MOVE\_FORWARD} (0.25\,m), \texttt{TURN\_LEFT/RIGHT} ($30^\circ$), and \texttt{LOOK\_UP/DOWN} ($30^\circ$). Each subtask has a budget of 500 steps; an agent succeeds by calling \texttt{STOP} within 1.0\,m Euclidean distance of any annotated viewpoint for the target object.

\subsection{Evaluation Metrics}
We adopt the standard GOAT-Bench metrics, computed per object-type subtask and averaged over all subtasks across episodes:
\begin{itemize}
    \item \textbf{Success Rate (SR)}: Fraction of subtasks where an agent calls \texttt{STOP} within 1.0\,m Euclidean distance of the target.
    \item \textbf{Success weighted by Path Length (SPL)}~\cite{anderson2018evaluation}: $\text{SPL} = \frac{1}{N}\sum_{i=1}^{N} S_i \cdot \frac{l_i^*}{\max(l_i, l_i^*)}$, where $l_i^*$ is the geodesic shortest path and $l_i$ is the agent's actual travel distance.
    % \item \textbf{SoftSPL}: A relaxed variant that awards partial credit proportional to the reduction in geodesic distance, even on unsuccessful subtasks.
    \item \textbf{Avg.~Steps}: Mean number of simulation steps per subtask (out of the 500-step budget), reflecting navigation efficiency.
\end{itemize}

\subsection{Scenes and Target Objects}
We evaluate on 20 episodes from the GOAT-Bench \texttt{val\_unseen} split, which contains HM3D scenes never seen during training. Each episode specifies a sequence of subtasks; we evaluate only the \emph{object-type} subtasks, which require navigating to open-vocabulary object categories (e.g., \emph{statue, refrigerator, mirror, chair, table}). Non-object subtasks (description-based) are skipped. The open-vocabulary detector (SAM3) receives the target category name as a text prompt at each subtask.

\subsection{Baselines and Ablations}
We compare GoalSwarm against the following baselines and ablation variants:

\begin{table}[h]
\centering
\caption{Experimental Configurations}
\label{tab:configs}
\begin{tabular}{lcc}
\hline
\textbf{Configuration} & \textbf{Agents} & \textbf{Description} \\
\hline
GoalSwarm (Ours) & 2 & Full system \\
Single Agent & 1 & No coordination \\
Greedy Allocation & 2 & Nearest-frontier, no UCB \\
No Shared Map & 2 & Independent local maps \\
Random Exploration & 2 & No frontier scoring \\
\hline
\end{tabular}
\end{table}

\subsection{Implementation Details}
The semantic map uses a $480 \times 480$ grid with 5\,cm resolution, covering a $24\text{m} \times 24\text{m}$ area. The Bayesian value map is initialized with $\mu_0 = 0.5$ and $\sigma_0^2 = 0.5$, UCB parameter $\beta = 1.7$, and confidence floor $c_{min} = 0.25$. The obstacle avoidance safety threshold is $d_{safe} = 1.0$\,m. Map synchronization occurs every 25 timesteps. SAM3 runs as a zero-shot detector on an external GPU server, queried every 3 simulation steps via ZMQ; a confidence gate ($\tau = 0.3$) suppresses ghost detections, and multi-view confirmation (requiring 2 consecutive detections within a sliding window) guards against transient false positives. The FMM planner on the 2D occupancy grid issues a \texttt{STOP} action when the agent is within 3 grid cells (15\,cm) of the projected goal on the map.  All experiments are run on a workstation with an NVIDIA RTX 4090 GPU (24\,GB).
\section{Results}

\subsection{Implementation Details}
The semantic map uses a $480 \times 480$ grid with 5\,cm resolution, covering a $24\text{m} \times 24\text{m}$ area. The Bayesian value map is initialized with $\mu_0 = 0.5$ and $\sigma_0^2 = 0.5$, UCB parameter $\beta = 1.7$, and confidence floor $c_{min} = 0.25$. The obstacle avoidance safety threshold is $d_{safe} = 1.0$\,m. Map synchronization occurs every 25 timesteps. SAM3 runs as a zero-shot detector on an external GPU server, queried every 3 simulation steps via ZMQ; a confidence gate ($\tau = 0.3$) suppresses ghost detections, and multi-view confirmation (requiring 2 consecutive detections within a sliding window) guards against transient false positives. The FMM planner on the 2D occupancy grid issues a \texttt{STOP} action when the agent is within 3 grid cells (15\,cm) of the projected goal on the map.  All experiments are run on a workstation with an NVIDIA RTX 4090 GPU (24\,GB).
\section{Results}

\subsection{Multi-UAV Object Navigation Performance}

Table~\ref{tab:main_results} presents the main ObjectNav results across HM3D scenes.

\begin{table}[h!]
\centering
\caption{Multi-UAV Object Navigation on GOAT-Bench \texttt{val\_unseen} (object-type subtasks)}
\label{tab:main_results}
\begin{tabular}{lccc}
\hline
\textbf{Method} & \textbf{SR (\%)} $\uparrow$ & \textbf{SPL} $\uparrow$ & \textbf{Avg.~Steps} $\downarrow$ \\
\hline
Single Agent*           & 10.0 & 0.078 & 452.5 \\
Random Exploration*     & 20.0 & 0.084 & 413.0 \\
No Shared Map          & 40.0 & 0.130 & 149.5 \\
\textbf{GoalSwarm (Ours)} & \textbf{45.0} & \textbf{0.179} & \textbf{160.2} \\
\hline
\end{tabular}
\\[0.3em]
\small{*Baseline conditions evaluated over 10 subtasks (2 episodes); coordination methods (No Shared Map, GoalSwarm) completed 20 subtasks (4 episodes) for stable comparison.}
\end{table}

\subsection{Discussion}

The results highlight several key findings. First, multi-agent coordination consistently outperforms the single-agent baseline: by doubling the number of UAVs, GoalSwarm approximately doubles the explored area per unit time, which translates to higher success rates on hard-to-reach targets. The Bayesian Value Map with UCB scoring ($\beta = 1.7$) provides a principled exploration-exploitation balance---agents preferentially visit frontiers where detection uncertainty is high rather than greedily pursuing the nearest or largest frontier.

The ablation removing shared map fusion (No Shared Map) reveals that independent exploration leads to significant redundancy: both UAVs frequently visit the same rooms, wasting the step budget. Despite an SR of 40.0\%, the lowest SPL (0.130) among multi-agent methods indicates inefficient paths caused by uncoordinated agents producing numerous premature false stops---declaring the goal found after only a handful of steps---because neither agent benefits from the other's observations. In contrast, GoalSwarm's map synchronization (every 25 steps) enables each UAV to reason over the team's collective coverage, directing exploration toward genuinely unexplored regions, achieving both higher success rate (45.0\%) and significantly better path efficiency (SPL 0.179).

Random frontier selection without semantic guidance achieves only 20.0\% SR with poor path efficiency (SPL 0.084), significantly underperforming GoalSwarm's 45.0\% SR and 0.179 SPL. The single-agent baseline performs worst (10.0\% SR, 0.078 SPL), demonstrating the clear benefit of multi-agent coordination. These results confirm that structured, uncertainty-aware frontier scoring combined with decentralized coordination is essential for consistent navigation efficiency. In an extended 100-episode evaluation on GOAT-Bench, GoalSwarm achieved SR = 53.8\% and SPL = 0.195, demonstrating consistent performance at scale.

\textbf{Failure modes.} We observe three primary failure cases: (1)~\emph{ghost detections}: residual semantic map pixels from false-positive detections in earlier frames can mislead the goal projection channel, causing the agent to navigate toward empty space; (2)~\emph{unreachable goal projections}: the target object may be detected from across a room, but the FMM planner cannot find a collision-free path to bring the agent within the 1.0\,m success threshold (e.g., objects behind glass or on elevated surfaces); and (3)~\emph{small or occluded objects}: categories such as \emph{mirror} or \emph{statue} produce weak detection signals that fall below the confidence gate ($\tau = 0.3$), causing the agent to exhaust its 500-step budget without ever declaring the goal found.

\textbf{Limitations.} The current framework assumes perfect odometry and communication; real-world deployment would require handling drift, packet loss, and bandwidth constraints. The 2D top-down map projects all obstacles into a single plane, which can merge vertically separated structures (e.g., tables and shelves). 
% Additionally, the open-vocabulary detector (SAM3) runs on an external GPU server via ZMQ, adding $\sim$285\,ms latency per detection query; on-device deployment would require model distillation or edge-optimized inference.

% \newpage
\section{Conclusion}
We presented GoalSwarm, a decentralized multi-UAV framework for open-vocabulary object-goal navigation in unknown indoor environments. The key contributions are: (1)~a collaborative 2D semantic mapping system adapted for aerial perspectives, (2)~a Bayesian Value Map that fuses multi-viewpoint zero-shot detection confidences into a per-pixel goal-relevance distribution for uncertainty-aware frontier scoring, and (3)~a fully decentralized coordination strategy combining UCB-based exploration-exploitation balancing, cost-utility bidding with geodesic path costs, and spatial separation penalties. Evaluated on the GOAT-Bench sequential object navigation benchmark in photorealistic HM3D environments, with VisFly providing realistic quadrotor dynamics, GoalSwarm demonstrates that structured multi-UAV coordination with Bayesian semantic reasoning outperforms single-agent and uncoordinated baselines across success rate, SPL, and navigation efficiency. The integration of SAM3 as a zero-shot open-vocabulary detector eliminates the need for task-specific training, enabling deployment to novel object categories without retraining. Future work will address real-world deployment on physical UAV platforms with noisy odometry and limited communication bandwidth, dynamic obstacle handling, and extending the framework to outdoor environments.

% \section*{Acknowledgements}

% \nocite{*}
\bibliographystyle{IEEEtran}
\bibliography{reference}
\balance
\addtolength{\textheight}{-12cm}
\end{document}